\documentclass[11pt,a4paper]{article}
\usepackage[hyperref]{acl2020}
\usepackage{times}
\usepackage{latexsym}

\usepackage{booktabs}
\usepackage{microtype}
\usepackage{times}
\usepackage{fancyhdr,graphicx,amsmath,amssymb}
\usepackage[ruled]{algorithm2e}
\include{pythonlisting}
\usepackage{inconsolata} 

\aclfinalcopy %

\usepackage{graphicx} %
\usepackage{amsmath}
\usepackage[rightcaption]{sidecap}
\usepackage{wrapfig}

\title{Bootstrapping Named Entity Recognition in E-Commerce with Positive Unlabeled Learning}

\author{Hanchu Zhang\textsuperscript{1} ~~~~~ Leonhard Hennig\textsuperscript{2} ~~~~~ Christoph Alt\textsuperscript{2} ~~~~~ Changjian Hu\textsuperscript{1} \\
\textbf{Yao Meng\textsuperscript{1} ~~~~~~~ Chao Wang\textsuperscript{1}}\\
\mbox{}\\
\textsuperscript{1}Lenovo Research Artificial Intelligence Lab  \\ \textsuperscript{2}German Research Center for Artificial Intelligence (DFKI)\\ 
\texttt{\{zhanghc9,hucj1,mengyao1,wangchao31\}@lenovo.com} \\
\texttt{\{leonhard.hennig,christoph.alt\}@dfki.de}}

\date{}

\begin{document}
\maketitle

\begin{abstract}
    Named Entity Recognition (NER) in domains like e-commerce is an understudied problem due to the lack of annotated datasets. Recognizing novel entity types in this domain, such as products, components, and attributes, is challenging because of their linguistic complexity and the low coverage of existing knowledge resources. To address this problem, we present a bootstrapped positive-unlabeled learning algorithm that integrates domain-specific linguistic features to quickly and efficiently expand the seed dictionary. The model achieves an average F1 score of $72.02\%$ on a novel dataset of product descriptions, an improvement of $3.63\%$ over a baseline BiLSTM classifier, and in particular exhibits better recall ($4.96\%$ on average). %
\end{abstract}

\section{Introduction}
    The vast majority of existing named entity recognition (NER) methods focus on a small set of prominent entity types, such as persons, organizations, diseases, and genes, for which labeled datasets are readily available~\cite{tjongkimsang2003conll,Smith2008OverviewOB,weischedel-etal-2011-ontonotes,Li2016BioCreativeVC}. There is a marked lack of studies in many other domains, such as e-commerce, and for novel entity types, e.g.\ products and components. %
    
    The lack of annotated datasets in the e-commerce domain makes it hard to apply supervised NER methods. An alternative approach is to use dictionaries~\cite{nadeau-etal-2006-unsupervised,yang-etal-2018-distantly}, but freely available knowledge resources, e.g.\ Wikidata~\cite{vrandecic-etal-2014-wikidata} or YAGO~\cite{suchanek-etal-2007-yago}, contain only very limited information about e-commerce entities. Manually creating a dictionary of sufficient quality and coverage would be prohibitively expensive. 
    This is amplified by the fact that in the e-commerce domain, entities are frequently expressed as complex noun phrases instead of proper names. Product and component category terms are often combined with brand names, model numbers, and attributes (\emph{``hard drive''} $\rightarrow$ \emph{``SSD hard drive''} $\rightarrow$ \emph{``WD Blue 500 GB SSD hard drive''}), which are almost impossible to enumerate exhaustively. 
    In such a low-coverage setting, employing a simple dictionary-based approach would result in very low recall, and yield very noisy labels when used as a source of labels for a supervised machine learning algorithm. To address the drawbacks of dictionary-based labeling, \citet{peng-etal-2019-distantly} propose a positive-unlabeled (PU) NER approach that labels positive instances using a seed dictionary, but makes no label assumptions for the remaining tokens~\cite{bekker-2018-pu-survey}. %
    The authors validate their approach on the CoNLL, MUC and Twitter datasets for standard entity types, but it is unclear how their approach transfers to the e-commerce domain and its entity types. %

    \paragraph{Contributions} We adopt the PU algorithm of \citet{peng-etal-2019-distantly} to the domain of consumer electronic product descriptions, and evaluate its effectiveness on four entity types: \emph{Product}, \emph{Component}, \emph{Brand} and \emph{Attribute}. Our algorithm bootstraps NER with a seed dictionary, iteratively labels more data and expands the dictionary, while accounting for accumulated errors from model predictions. %
    During labeling, we utilize dependency parsing to efficiently expand dictionary matches in text.
    Our experiments on a novel dataset of product descriptions show that this labeling mechanism, combined with a PU learning strategy, consistently improves F1 scores over a standard BiLSTM classifier. Iterative learning quickly expands the dictionary, and further improves model performance. The proposed approach exhibits much better recall than the baseline model, and generalizes better to unseen entities.

\section{NER with Positive Unlabeled Learning}
\label{sec:pu-ner}
    In this section, we first describe the iterative bootstrapping process, followed by our approach to positive unlabeled learning for NER (PU-NER).
    
    \subsection{Iterative Bootstrapping}
    \label{sec:iter-bootstrap}
    
    \begin{algorithm}[t!]
            \KwIn{Dictionary $D_{seed}$, Corpus $C$, threshold $K$, max\_iterations $I$}
            \KwResult{Dictionary $D^+$, Classifier $L$}
             $D^+ \leftarrow D_{seed}$\;
             $C_{dep} \leftarrow dependency\_parse(C$)\;
             $i \leftarrow 0$\;
             \While{not\_converged($D^+$) and i $<$ I}{ 
              $C_{lab} \leftarrow label(C,D^+)$\;
              $C_{exp} \leftarrow expand\_labels(C_{lab},C_{dep})$\;
              $L \leftarrow train\_classifier(C_{exp})$\;
              $C_{pred} \leftarrow predict(C_{exp}, L)$\;
              \For{$e \leftarrow C_{pred}$}{
                \If{e $\notin D^+$ and freq(e) $>$ K}{
                    $D^+ \leftarrow add\_entity(D^+, e)$\;
                }
              }
              $i \leftarrow i + 1$\;
            }
           
             \caption{Iterative Bootstrapping NER}
              \label{alg:iterative-bootstrapping}
        \end{algorithm}
    
\begin{figure}[t!]
            \includegraphics[width=0.48\textwidth]{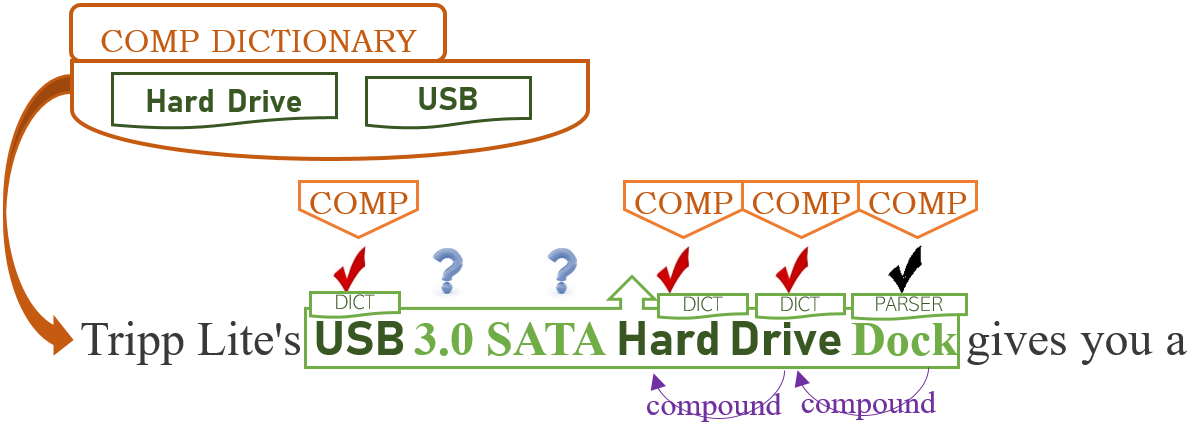}
            \caption{Red check marks indicate tokens labeled by the dictionary, black those based on label expansion using dependency information. The green box shows the true extent of the multi-token \emph{Component} entity.}
            \label{fig:dictionary-labeling}
        \end{figure}

        The goal of iterative bootstrapping is to successively expand a seed dictionary of entities to label an existing training dataset, improving the quality and coverage of labels in each iteration
         (see Algorithm~\ref{alg:iterative-bootstrapping}).
        In the first step, we use the seed dictionary to assign initial labels to each token. 
        We then utilize the dependency parses of sentences to label tokens in a ``compound'' relation with already labeled tokens (see Figure~\ref{fig:dictionary-labeling}). 
        In the example ``hard drive'' is labeled a \emph{Component} based on the initial seed dictionary, and according to its dependency parse it has a ``compound'' relation with ``dock'', which is therefore also labeled as a \emph{Component}. We employ an IO label scheme, because dictionary entries are often more generic than the specific matches in text (see the previous example), which would lead to erroneous tags with schemes such as BIO. %
        
        In the second step, we train a NER model on the training dataset with new labels assigned. We repeat these steps at most $I$ times, and in each subsequent iteration we use the trained model to predict new token-level labels on the training data. Novel entities predicted more than $K$ times are included in the dictionary for the next labeling step. The threshold $K$ ensures that we do not introduce noise in the dictionary with spurious positively labeled entities.

    \subsection{PU-NER Model}
    \label{sec:pu-model}
\begin{figure}[t!]
             \includegraphics[width=0.48\textwidth]{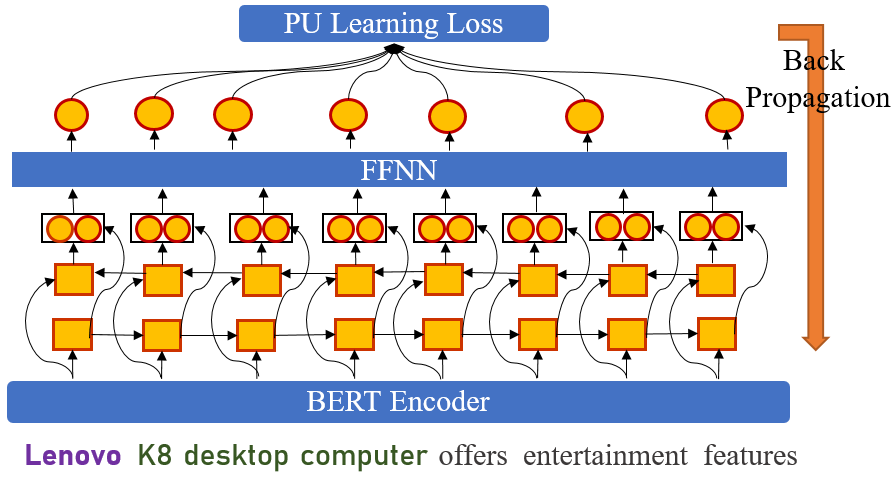}
             \caption{Architecture of the positive unlabeled NER (PU-NER) model.}
             \label{fig:architecture}
             \vspace{-10pt}
         \end{figure}

        As shown in Figure~\ref{fig:architecture},
        our model first uses BERT~\cite{devlin-2018-bert} to encode the sub-word tokenized input text into a sequence of contextualized token representations ${\{z_1,...,z_L\}}$, followed by a bidirectional LSTM~\citep{lample-etal-2016-neural} layer to model further interactions between tokens. %
        Similar to \citet{devlin-2018-bert}, we treat NER as a token-level classification task, without using a CRF to model dependencies between entity labels. We use the vector associated with the first sub-word token in each word as the input to the entity classifier, which consists of a feedforward neural network with a single projection layer. We use back propagation to update the training parameters of the Bi-LSTM and the final classifier, without fine-tuning the entire BERT model.  
        
        Dictionary-based labeling achieves high precision on the matched entities but low recall. This fits the positive unlabeled setting \citep{elkan-etal-2008-learning}, which assumes that a learner only has access to positive examples and unlabeled data.
        Thus, we consider all tokens matched by the dictionary as positive, and consider all other tokens to be unlabeled.
        The goal of PU learning is then to estimate the true risk regarding the expected number of positive examples remaining in the unlabeled data. We define the empirical risk as $\hat{R_l}=\frac{1}{n}\sum_{i}^{n}l(\hat{y_i},y_i)$ and assume the class prior to be equal to real distribution of examples in the data $\pi_p=P(Y=1)$, and $\pi_n=P(Y=0)$. As the model tends to predict the positive labels correctly during training, i.e.\ $l(\hat{y_i}^p,1)$ declines to a small value. We follow \citet{peng-etal-2019-distantly} and combine risk estimation with a non-negative constraint:
        \begin{equation*}
            \begin{split}
                &\hat{R_l}=\frac{1}{n_p}\sum_{i}^{n_p}l(\hat{y_i}^p,1) \\
                &+max\left(0,\frac{1}{n_u}\sum_{i}^{n_u}l(\hat{y_i}^u,0)-\frac{\pi_p}{n_p}\sum_{i}^{n_p}l(\hat{y_i}^p,0)\right) \\
            \end{split}
        \end{equation*}

\section{Dataset}
\label{sec:dataset}
    E-commerce covers a wide range of complex entity types. In this work, we focus on electronic products, e.g.\ personal computers, mobile phones, and related hardware, and define the following entity types: \textbf{Products}, i.e.\ electronic consumer devices such as mobiles, laptops, and PCs. \emph{Products} may be preceded by a brand and include some form of model, year, or version specification, e.g.\ ``Galaxy S8'' or ``Dell Latitude 6400 multimedia notebook''.
    \textbf{Components} are parts of a product, typically with a physical aspect, e.g.\ ``battery'', or ``multimedia keyboard''.\footnote{Non-physical product features and software, such as ``Toshiba Face Recognition Software'', or ``Windows 7'' are not considered as components.} \textbf{Brand} refers to producers of a \emph{product} or \emph{component}, e.g.\ "Samsung", or "Dell". \textbf{Attributes} are units associated with components, e.g.\ size (``4 TB''), or weight (``3 kg''). %
    
    To create our evaluation dataset, we use the \emph{Amazon review dataset}~\citep{McAuley2015ImageBasedRO},\footnote{http://jmcauley.ucsd.edu/data/amazon/links.html} a collection of product metadata and customer reviews from Amazon. The metadata includes product title, a descriptive text, category information, price, brand, and image features.
    We use only entries in the \emph{Electronics/Computers} subcategory and randomly sample product descriptions of length 500--1000 characters, yielding a dataset of 24,272 training documents. We randomly select another 100 product descriptions to form the final test set. These are manually annotated by 2 trained linguists, with disagreements resolved by a third expert annotator. Token-level inter-annotator agreement was high (Krippendorf's $\alpha = 0.7742$). The test documents contain a total of $27,108$ tokens ($1,493$ \emph{Product}, $3,234$ \emph{Component}, $1,485$ \emph{Attribute}, and $443$ \emph{Brand}).

\section{Experiments}
\label{sec:experiments}
    To evaluate our proposed model (\emph{PU}), we compare it against two baselines: (1) dictionary-only labeling (\emph{Dictionary}), and (2) our model with standard cross-entropy loss instead of the PU learning risk (\emph{BiLSTM}). The \emph{BiLSTM} model is trained in a supervised fashion, treating all non-dictionary entries as negative tokens. The \emph{BiLSTM} and \emph{PU} models were implemented using AllenNLP~\citep{gardner-etal-2018-allennlp}. We use SpaCy\footnote{https://spacy.io/} for preprocessing, dependency parsing, and dictionary-based entity labeling. We manually define seed dictionaries for \emph{Product} (6 entries), \emph{Component} (60 entries) and \emph{Brand} (13 entries). For \emph{Attribute}s, we define a set of 8 regular expressions to pre-label the dataset. Following previous works, we evaluate model performance using token-level F1 score. 
    
   There are two options to estimate the value of the class prior $\pi_p$. One approach is to treat $\pi_p$ as a hyperparameter which is fixed during training. Another option is suggested in \citet{peng-etal-2019-distantly}, who specify an initial value for $\pi_p$ to start bootstrapping, but recalculate $\pi_p$ after several train-relabel steps based on the predicted entity type distribution. In our work, we treated $\pi_p$ as a fixed hyperparameter with a value of $\pi_p=0.01$.
  
    We run our bootstrapping approach for $I=10$ iterations, and report the F1 score of the best iteration.
    \begin{table*}[t!]
        \centering
        \begin{tabular}{lcccccc}
            \toprule
            Entity Type & Dictionary & BiLSTM & PU & PU+Dep & PU+Iter & PU+Dep+Iter \\
            \midrule
            Component  & 46.19 & 65.98  & 66.89  & \ 67.38  & 68.67 & \textbf{70.66}   \\
            Product  & 16.78 & 60.23  & 60.24 & 65.05 & 60.24 & \textbf{67.07}   \\
            Brand  & 49.74 & 74.06 & 74.84 & \textbf{76.24}  & \textbf{76.24}  & \textbf{76.24}   \\
            Attribute  & 7.05  & 73.30 & 73.84 & \textbf{74.14}  & \textbf{74.14} & \textbf{74.14}  \\
            
            \midrule
            All & 29.94 & 68.39 & 68.95 & 70.70& 69.82& \textbf{72.02} \\
            \bottomrule
        \end{tabular}
        \caption{
        Token-Level F1 scores on the test set. The unmodified PU algorithm achieves an average F1 score of $68.95\%$. Integrating dependency parsing (Dep) and iterative relabeling (Iter) raises the F1 score to $72.02\%$, an improvement of $42.08\%$ over a dictionary-only approach, and $3.63\%$ over a BiLSTM baseline.}
        \label{tab:results_main}
        \vspace{-10pt}
    \end{table*}
    \subsection{Results and Discussion}
    \label{sec:results}
\begin{figure}[ht!]
    \includegraphics[width=\columnwidth,trim=0 5 40 30,clip]{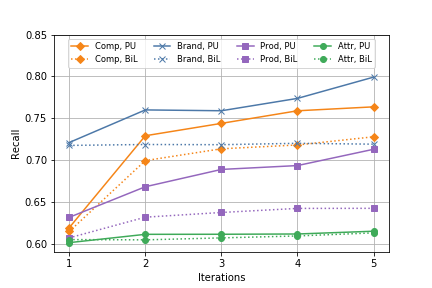}
    \caption{Recall curves of the BiLSTM+Dep and PU+Dep model for \emph{Component}, \emph{Product}, \emph{Brand}, and \emph{Attribute}. PU+Dep boosts recall by $3.03\%$ on average, with a max average difference of $4.96\%$ after 5 iterations.}
    \label{fig:iterations_vs_recall}
    \end{figure}

 Table~\ref{tab:results_main} shows the F1 scores of several model ablations by entity type on our test dataset. From the table, we can observe: 1) The PU algorithm outperforms the simpler models for most classes, which demonstrates the effectiveness of the PU learning framework for NER in our domain. 2)
    Dependency parsing is a very effective feature for \emph{Component} and \emph{Product}, and it strongly improves the overall F1 score. 3) The iterative training strategy yields a significant improvement for most classes. Even after several iterations, it still finds new entries to expand the dictionaries (Figure~\ref{fig:iterations_vs_recall}).
    
    The \emph{Dictionary} approach shows poor performance on average, which is due to the low recall caused by very limited entities in the dictionary. \emph{PU} greatly outperforms the dictionary approach, and has an edge in F1 score over the \emph{BiLSTM} model. The advantages of PU gradually accumulate with each iteration. For \emph{Product}, the combination of PU learning, dependency parsing-based labeling, and iterative bootstrapping, yields a $7\%$ improvement in F1 score, for \emph{Component}, it is still $5\%$.
    \paragraph{PU Learning Performance}
    Figure~\ref{fig:iterations_vs_recall} shows that the \emph{PU} algorithm especially improves recall over the baseline classifier for \emph{Components}, \emph{Products} and \emph{Brands}.
    With each iteration step, the PU model is increasingly better able to predict unseen entities, and achieves higher recall scores than the BiLSTM model. While the baseline curve on \emph{Brands} stays almost flat during iterations, \emph{PU} consistently improves recall as new entities are added into dictionary. For \emph{Attributes}, however, both models exhibit about the same level of recall, which in addition is largely unaffected by the number of iterations. %
   
   This suggests that PU learning better estimates the true loss in the model. In a fully supervised setting, a standard classification loss function can accurately describe the loss on positive and negative samples. However, in the positive unlabeled setting, many unlabeled samples may actually be positive, and therefore the computed loss should not strongly push the model towards the negative class. We therefore want to quantify how much the loss is overestimated due to false negative samples, so that we can appropriately reduce this loss using the estimated real class distribution.
   
    \paragraph{Error Analysis}
    Both PU and the baseline model in some cases have difficulties predicting \emph{Attribute}s correctly. This can be due to spelling differences between train and test data (e.g.\ "8 Mhz" vs "8Mhz"), but also because of unclean texts in the source documents. Another source of errors is the fixed word piece vocabulary of the pre-trained BERT model, which often splits unit terms such as "Mhz" into several word pieces. Since we use only the first word piece of a token for prediction, this means that signals important for prediction of the  \emph{Attribute} class may get lost. This suggests that for technical domains with very specific vocabulary, tokenization is important to allow the model to better represent the meaning of each word piece.

\section{Related work}
\label{sec:rel-work}

Recent work in positive-unlabeled learning in the area of NLP includes deceptive review detection~\cite{ren-etal-2014-positive}, keyphrase extraction~\cite{sterckx-etal-2016-supervised} and fact check-worthiness detection~\cite{Wright2020FactCD}, see also \cite{bekker-2018-pu-survey} for a survey. Our approach extends the work of~\citet{peng-etal-2019-distantly} in a novel domain and for challenging entity types. %
In the area of NER for e-commerce,  \citet{putthividhya-hu-2011-bootstrapped} present an approach to extract product attributes and values from product listing titles. \citet{Zheng2018OpenTagOA} formulate missing attribute value extraction as a sequence tagging problem, and present a BiLSTM-CRF model with attention. \citet{Pazhouhi2018AutomaticPN} studies the problem of product name recognition, but uses a fully supervised approach. In contrast, our method is semi-supervised and uses only very few seed labels. %
\section{Conclusion}
In this work, we introduce a bootstrapped, iterative NER model that integrates a PU learning algorithm for recognizing named entities in a low-resource setting. Our approach combines dictionary-based labeling with syntactically-informed label expansion to efficiently enrich the seed dictionaries. Experimental results on a dataset of manually annotated e-commerce product descriptions demonstrate the effectiveness of the proposed framework.

\section*{Acknowledgments}
We would like to thank the reviewers for their valuable comments and feedback. Christoph Alt and Leonhard Hennig have been partially supported by the German Federal Ministry for Economic Affairs and Energy as part of the project PLASS (01MD19003E).

\bibliography{main}

\begin{thebibliography}{21}
\expandafter\ifx\csname natexlab\endcsname\relax\def\natexlab#1{#1}\fi

\bibitem[{Bekker and Davis(2018)}]{bekker-2018-pu-survey}
Jessa Bekker and Jesse Davis. 2018.
\newblock \href {http://arxiv.org/abs/1811.04820} {Learning from positive and
  unlabeled data: {A} survey}.
\newblock \emph{CoRR}, abs/1811.04820.

\bibitem[{Devlin et~al.(2018)Devlin, Chang, Lee, and
  Toutanova}]{devlin-2018-bert}
Jacob Devlin, Ming{-}Wei Chang, Kenton Lee, and Kristina Toutanova. 2018.
\newblock \href {http://arxiv.org/abs/1810.04805} {{BERT:} pre-training of deep
  bidirectional transformers for language understanding}.
\newblock \emph{CoRR}, abs/1810.04805.

\bibitem[{Elkan and Noto(2008)}]{elkan-etal-2008-learning}
Charles Elkan and Keith Noto. 2008.
\newblock Learning classifiers from only positive and unlabeled data.
\newblock In \emph{Proceedings of the 14th ACM SIGKDD international conference
  on Knowledge discovery and data mining}, pages 213--220.

\bibitem[{Gardner et~al.(2018)Gardner, Grus, Neumann, Tafjord, Dasigi, Liu,
  Peters, Schmitz, and Zettlemoyer}]{gardner-etal-2018-allennlp}
Matt Gardner, Joel Grus, Mark Neumann, Oyvind Tafjord, Pradeep Dasigi,
  Nelson~F. Liu, Matthew Peters, Michael Schmitz, and Luke Zettlemoyer. 2018.
\newblock \href {https://doi.org/10.18653/v1/W18-2501} {{A}llen{NLP}: A deep
  semantic natural language processing platform}.
\newblock In \emph{Proceedings of Workshop for {NLP} Open Source Software
  ({NLP}-{OSS})}, pages 1--6, Melbourne, Australia. Association for
  Computational Linguistics.

\bibitem[{Lample et~al.(2016)Lample, Ballesteros, Subramanian, Kawakami, and
  Dyer}]{lample-etal-2016-neural}
Guillaume Lample, Miguel Ballesteros, Sandeep Subramanian, Kazuya Kawakami, and
  Chris Dyer. 2016.
\newblock \href {https://doi.org/10.18653/v1/N16-1030} {Neural architectures
  for named entity recognition}.
\newblock In \emph{Proceedings of the 2016 Conference of the North {A}merican
  Chapter of the Association for Computational Linguistics: Human Language
  Technologies}, pages 260--270, San Diego, California. Association for
  Computational Linguistics.

\bibitem[{Li et~al.(2016)Li, Sun, Johnson, Sciaky, Wei, Leaman, Davis,
  Mattingly, Wiegers, and Lu}]{Li2016BioCreativeVC}
Jiao Li, Yueping Sun, Robin~J. Johnson, Daniela Sciaky, Chih-Hsuan Wei, Robert
  Leaman, Allan~Peter Davis, Carolyn~J. Mattingly, Thomas~C. Wiegers, and
  Zhiyong Lu. 2016.
\newblock Biocreative v cdr task corpus: a resource for chemical disease
  relation extraction.
\newblock \emph{Database : the journal of biological databases and curation},
  2016.

\bibitem[{McAuley et~al.(2015)McAuley, Targett, Shi, and van~den
  Hengel}]{McAuley2015ImageBasedRO}
Julian~J. McAuley, Christopher Targett, Qinfeng Shi, and Anton van~den Hengel.
  2015.
\newblock Image-based recommendations on styles and substitutes.
\newblock \emph{Proceedings of the 38th International ACM SIGIR Conference on
  Research and Development in Information Retrieval}.

\bibitem[{Nadeau et~al.(2006)Nadeau, Turney, and
  Matwin}]{nadeau-etal-2006-unsupervised}
David Nadeau, Peter~D. Turney, and Stan Matwin. 2006.
\newblock Unsupervised named-entity recognition: Generating gazetteers and
  resolving ambiguity.
\newblock In \emph{Advances in Artificial Intelligence}, pages 266--277,
  Berlin, Heidelberg. Springer Berlin Heidelberg.

\bibitem[{Pazhouhi(2018)}]{Pazhouhi2018AutomaticPN}
Elnaz Pazhouhi. 2018.
\newblock Automatic product name recognition from short product descriptions.
\newblock Master's thesis, University of Twente.

\bibitem[{Peng et~al.(2019)Peng, Xing, Zhang, Fu, and
  Huang}]{peng-etal-2019-distantly}
Minlong Peng, Xiaoyu Xing, Qi~Zhang, Jinlan Fu, and Xuanjing Huang. 2019.
\newblock \href {https://doi.org/10.18653/v1/P19-1231} {Distantly supervised
  named entity recognition using positive-unlabeled learning}.
\newblock In \emph{Proceedings of the 57th Annual Meeting of the Association
  for Computational Linguistics}, pages 2409--2419, Florence, Italy.
  Association for Computational Linguistics.

\bibitem[{Putthividhya and Hu(2011)}]{putthividhya-hu-2011-bootstrapped}
Duangmanee Putthividhya and Junling Hu. 2011.
\newblock \href {https://www.aclweb.org/anthology/D11-1144} {Bootstrapped named
  entity recognition for product attribute extraction}.
\newblock In \emph{Proceedings of the 2011 Conference on Empirical Methods in
  Natural Language Processing}, pages 1557--1567, Edinburgh, Scotland, UK.
  Association for Computational Linguistics.

\bibitem[{Ren et~al.(2014)Ren, Ji, and Zhang}]{ren-etal-2014-positive}
Yafeng Ren, Donghong Ji, and Hongbin Zhang. 2014.
\newblock \href {https://doi.org/10.3115/v1/D14-1055} {Positive unlabeled
  learning for deceptive reviews detection}.
\newblock In \emph{Proceedings of the 2014 Conference on Empirical Methods in
  Natural Language Processing ({EMNLP})}, pages 488--498, Doha, Qatar.
  Association for Computational Linguistics.

\bibitem[{Smith et~al.(2008)Smith, Tanabe, nee Ando, Kuo, Chung, Hsu, Lin,
  Klinger, Friedrich, Ganchev, Torii, Liu, Haddow, Struble, Povinelli, Vlachos,
  Baumgartner, Hunter, Carpenter, Tsai, Dai, Liu, Chen, Sun, Katrenko,
  Adriaans, Blaschke, Torres, Neves, Nakov, Divoli, Ma{\~n}a-L{\'o}pez, Mata,
  and Wilbur}]{Smith2008OverviewOB}
Larry Smith, Lorraine~K. Tanabe, Rie~Johnson nee Ando, Cheng-Ju Kuo, I-Fang
  Chung, Chun-Nan Hsu, Yu-Shi Lin, Roman Klinger, Christoph~M. Friedrich,
  Kuzman Ganchev, Manabu Torii, Hongfang Liu, Barry Haddow, Craig~A. Struble,
  Richard~J. Povinelli, Andreas Vlachos, William~A. Baumgartner, Lawrence~E.
  Hunter, Bob Carpenter, Richard Tzong-Han Tsai, Hong-Jie Dai, Feng Liu, Yifei
  Chen, Chengjie Sun, Sophia Katrenko, Pieter Adriaans, Christian Blaschke,
  Rafael Torres, Mariana Neves, Preslav Nakov, Anna Divoli, Manuel~J.
  Ma{\~n}a-L{\'o}pez, Jacinto Mata, and W.~John Wilbur. 2008.
\newblock Overview of biocreative ii gene mention recognition.
\newblock \emph{Genome Biology}, 9:S2 -- S2.

\bibitem[{Sterckx et~al.(2016)Sterckx, Caragea, Demeester, and
  Develder}]{sterckx-etal-2016-supervised}
Lucas Sterckx, Cornelia Caragea, Thomas Demeester, and Chris Develder. 2016.
\newblock \href {https://doi.org/10.18653/v1/D16-1198} {Supervised keyphrase
  extraction as positive unlabeled learning}.
\newblock In \emph{Proceedings of the 2016 Conference on Empirical Methods in
  Natural Language Processing}, pages 1924--1929, Austin, Texas. Association
  for Computational Linguistics.

\bibitem[{Suchanek et~al.(2007)Suchanek, Kasneci, and
  Weikum}]{suchanek-etal-2007-yago}
Fabian~M. Suchanek, Gjergji Kasneci, and Gerhard Weikum. 2007.
\newblock \href {https://doi.org/10.1145/1242572.1242667} {Yago: A core of
  semantic knowledge}.
\newblock In \emph{Proceedings of the 16th International Conference on World
  Wide Web}, WWW ’07, page 697–706, New York, NY, USA. Association for
  Computing Machinery.

\bibitem[{Tjong Kim~Sang and De~Meulder(2003)}]{tjongkimsang2003conll}
Erik~F. Tjong Kim~Sang and Fien De~Meulder. 2003.
\newblock Introduction to the conll-2003 shared task: Language-independent
  named entity recognition.
\newblock In \emph{Proceedings of CoNLL-2003}, pages 142--147. Edmonton,
  Canada.

\bibitem[{Vrande\v{c}ic and Kr\"{o}tzsch(2014)}]{vrandecic-etal-2014-wikidata}
Denny Vrande\v{c}ic and Markus Kr\"{o}tzsch. 2014.
\newblock \href {https://doi.org/10.1145/2629489} {Wikidata: A free
  collaborative knowledgebase}.
\newblock \emph{Commun. ACM}, 57(10):78–85.

\bibitem[{Weischedel et~al.(2011)Weischedel, Hovy, Marcus, Palmer, Belvin,
  Pradhan, Ramshaw, and Xue}]{weischedel-etal-2011-ontonotes}
Ralph Weischedel, Eduard Hovy, Mitchell Marcus, Martha Palmer, Robert Belvin,
  Sameer Pradhan, Lance Ramshaw, and Nianwen Xue. 2011.
\newblock \emph{OntoNotes: A Large Training Corpus for Enhanced Processing}.

\bibitem[{Wright and Augenstein(2020)}]{Wright2020FactCD}
Dustin Wright and Isabelle Augenstein. 2020.
\newblock Fact check-worthiness detection as positive unlabelled learning.
\newblock \emph{ArXiv}, abs/2003.02736.

\bibitem[{Yang et~al.(2018)Yang, Chen, Li, He, and
  Zhang}]{yang-etal-2018-distantly}
Yaosheng Yang, Wenliang Chen, Zhenghua Li, Zhengqiu He, and Min Zhang. 2018.
\newblock \href {https://www.aclweb.org/anthology/C18-1183} {Distantly
  supervised {NER} with partial annotation learning and reinforcement
  learning}.
\newblock In \emph{Proceedings of the 27th International Conference on
  Computational Linguistics}, pages 2159--2169, Santa Fe, New Mexico, USA.
  Association for Computational Linguistics.

\bibitem[{Zheng et~al.(2018)Zheng, Mukherjee, Dong, and
  Li}]{Zheng2018OpenTagOA}
Guineng Zheng, Subhabrata Mukherjee, Xin Dong, and Feifei Li. 2018.
\newblock Opentag: Open attribute value extraction from product profiles.
\newblock \emph{Proceedings of the 24th ACM SIGKDD International Conference on
  Knowledge Discovery \& Data Mining}.

\end{thebibliography}
\bibliographystyle{acl_natbib}

\appendix

\end{document}